\documentclass[preprint,12pt]{elsarticle}

\usepackage{amssymb}
\usepackage{amsmath}
\usepackage{subcaption}
\usepackage{booktabs}

\usepackage{color}
\definecolor{red}{rgb}{1,0,0}
\definecolor{blue}{rgb}{0,0,1}
\usepackage{ulem}

\begin{document}

\begin{frontmatter}



\title{RoIPoly: Vectorized Building Outline Extraction Using Vertex and Logit Embeddings}


\author[inst1]{Weiqin Jiao\corref{cor1}}

\affiliation[inst1]{organization={Department of Earth Observation Science, Faculty of Geo-Information Science and Earth Observation (ITC), University of Twente} %Department and Organization
            %addressline={Hallenweg 8}, 
            %city={Enschede},
            %postcode={7522 NH}, 
            %state={Overijssel},
            %country={The Netherlands}
            }
\cortext[cor1]{Corresponding author. Email: w.jiao@utwente.nl}
\author[inst1]{Hao Cheng}
\author[inst1]{George Vosselman}
\author[inst1]{Claudio Persello}

\begin{abstract}
Polygonal building outlines are crucial for geographic and cartographic applications. 
The existing approaches for outline extraction from aerial or satellite imagery are typically decomposed into subtasks, \textit{e.g.,} building masking and vectorization, or treat this task as a sequence-to-sequence prediction of ordered vertices. 
The former lacks efficiency, and the latter often generates redundant vertices, both resulting in suboptimal performance. 
To handle these issues, we propose a novel Region-of-Interest (RoI) query-based approach called RoIPoly. Specifically, we formulate each vertex as a query and constrain the query attention on the most relevant regions of a potential building, yielding reduced computational overhead and more efficient vertex-level interaction. 
Moreover, we introduce a novel learnable logit embedding to facilitate vertex classification on the attention map; 
thus, no post-processing is needed for redundant vertex removal. 
We evaluated our method on the vectorized building outline extraction dataset CrowdAI and the 2D floorplan reconstruction dataset Structured3D. 
On the CrowdAI dataset, RoIPoly with a ResNet50 backbone outperforms existing methods with the same or better backbones on most MS-COCO metrics, especially on small buildings, and achieves competitive results in polygon quality and vertex redundancy without any post-processing. 
On the Structured3D dataset, our method achieves the second-best performance on most metrics among existing methods dedicated to 2D floorplan reconstruction, demonstrating our cross-domain generalization capability. 
The code will be released upon acceptance of this paper.
\end{abstract}




\end{frontmatter}


\section{Introduction}
\label{sec:intro}
Vectorized building outline extraction refers to automatically predicting building polygons from aerial or satellite images, an important task with a wide range of geospatial applications, such as topographic \cite{li2019topological} and cadastral mapping \cite{10282644}, as well as disaster management \cite{lu2004change}. However, directly extracting an accurate building outline with a variable number of vertices is challenging. Different from object detection, it requires delineating buildings from an image and contouring the buildings of arbitrary shapes using an unknown number of connected vertices.

Generally, approaches to building outline extraction are divided into multi-step and end-to-end paradigms. In the multi-step paradigm, typically, images are first processed to generate segmentation masks, and then polygon extraction is performed on the intermediate representations. Despite promising performance, these methods \cite{girard2021polygonal,li2021joint,xu2023hisup} incur low efficiency during inference due to their reliance on post-processing methods to vectorize polygons and reduce redundant vertices, which heavily limits their practical application. On the other hand, end-to-end methods directly learn building polygons as ordered vertex sequences of different lengths using \textit{e.g.}, Recurrent Neural Networks (RNNs) and/or Transformers \cite{zorzi2022polyworld, hu2023polybuilding}. Due to the lack of efficient vertex-level interaction and various numbers of vertices in each polygon, many redundant vertices are generated, especially at building corners. To remove these redundant vertices, post-processing is often needed \cite{hu2023polybuilding}. In addition, existing methods are prone to generate intricate artifacts along edges, which significantly impacts their performance on small buildings.

\begin{figure}[tb]
  \centering
  \includegraphics[trim={0.cm 15cm 2.6cm 0.cm}, clip, width=\linewidth]{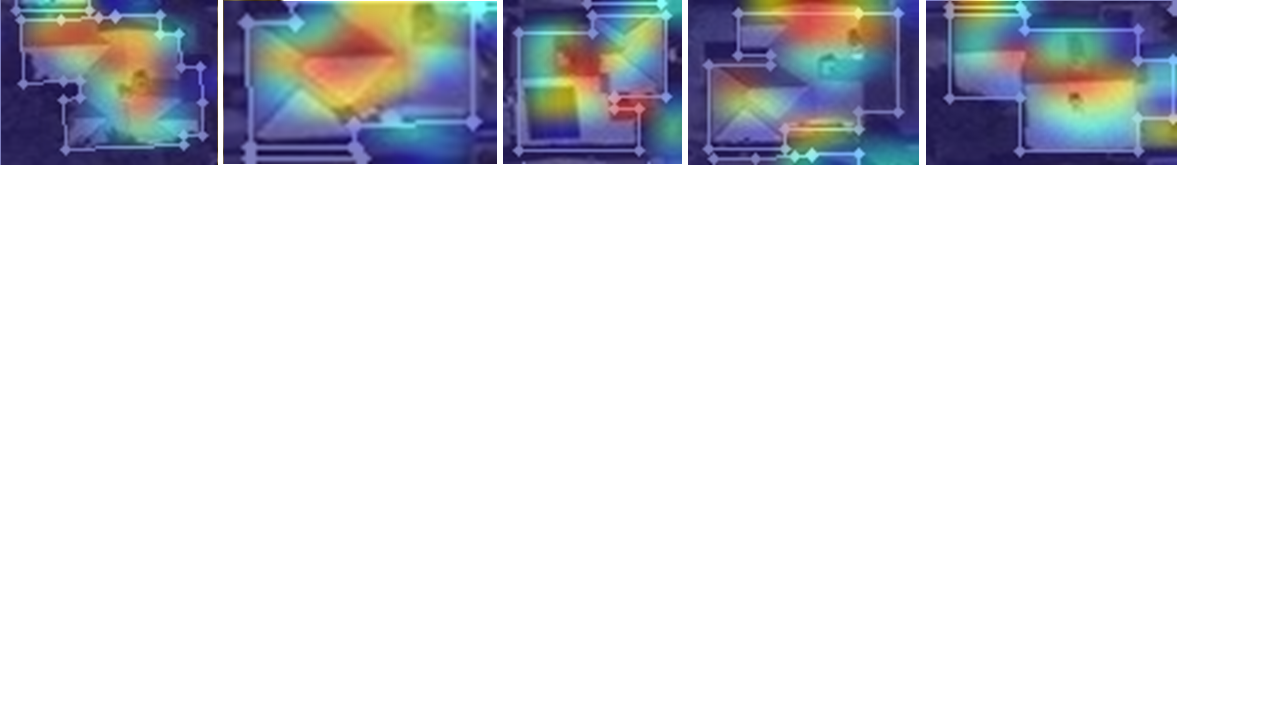}
  \caption{Examples of the Region-of-Interest (RoI) features of different buildings. A warmer color depicts a higher value in the RoI feature maps, indicating the visual cues for the buildings (highlighted in white polylines) to be outlined. Our RoI query-based approach ensures vertex query attention is confined to building RoIs so as to exclude irrelevant background features and improve attention efficiency. It should be noted that the polylines in this figure only serve as the visual reference to show the alignment between the buildings and feature maps; they are not included in the feature maps for the model training.
  \vspace{-6pt}
  }
  \label{fig:roi_feature_viz_examples}
\end{figure}

Recently, a few attempts have explored query-based approaches to extract vectorized building outlines in an end-to-end manner. 
For example, PolyBuilding \cite{hu2023polybuilding} formulates each polygon as a query to enable polygon-level interaction across buildings. 
However, it lacks vertex-level interaction within each building, which is crucial to distinguish between valid and invalid vertices to avoid predicting redundant vertices. 
RoomFormer \cite{yue2023connecting}, proposed for room polygon extraction from 2D floorplans, formulates the vertices of all the polygons (\textit{i.e.}, rooms) in an image as queries, enabling vertex-level interaction of both inter and intra polygons. 
However, the complexity of this approach grows quadratically with the number of polygons and vertices. 
Hence, it is not suitable for building outline extraction as one image contains multiple buildings and each building requires many vertices to model its arbitrary shape, as opposed to mostly rectangular rooms. 

In this work, we introduce a novel query-based approach for vectorized building outline extraction. 
Following the end-to-end pipeline, we predict a fixed number of vertices for each polygon and a class label for each vertex to distinguish between valid and invalid vertices. 
The central idea is to facilitate efficient and effective vertex-level interaction to address two common issues in existing methods: 1) the need for post-processing to remove redundant vertices; 2) the tendency to produce intricate artifacts along edges, especially on small buildings. 
To achieve this, we formulate each vertex as a query and constrain vertex-level interaction to only relevant areas of each building, thereby minimizing computational costs while ensuring relevant feature pooling. 
This is particularly helpful for small buildings, as their relevant features only occupy a small region of the image and are easily affected by irrelevant surrounding features. 
As illustrated in Fig. \ref{fig:roi_feature_viz_examples}, we use RoIAlign \cite{he2017mask} to extract the Region-of-Interest (RoI) features for each building and confine the query attention within it. 
The geometrical structure of each building in the RoI is highlighted and distinguishable from the background features. 
Hence, the polygon-level interaction across buildings is not necessary, thus largely reducing computational costs. 
Moreover, inspired by the learnable positional embeddings in DETR-based models \cite{zhu2020deformable,liu2022dab} for feature pooling around relevant positions, we propose to leverage vertex classification logit as an additional learnable logit embedding to introduce classification information to the vertex-level interaction. 
Unlike the learnable positional embedding directly added to the vertex queries in the Transformer decoder \cite{zhu2020deformable, liu2022dab}, we adopt adaptive layer norm (adaLN) from diffusion models \cite{peebles2023scalable}, originally designed for processing conditional information, to fuse heterogeneous positional and logit embeddings into the vertex queries. 
The improved vertex classification performance allows us to remove redundant vertices according to their classification scores without any further post-processing. 

We evaluate our model on the mainstream dataset for vectorized building outline extraction, CrowdAI \cite{mohanty2020deep}, and show superior performance in terms of pixel coverage, vertex redundancy, and polygon quality, i.e., achieving 83.6 Average Precision (AP), exhibiting a 4.2 AP gain compared to the well-established multi-step model HiSup \cite{xu2023hisup}, and pushing the ${\rm AP_S}$ from 54.3\% to 74.2\% for small-sized building polygon extraction. 
We further extend our model to the challenging task of 2D floorplan reconstruction, which aims to reconstruct a 2D vector map in a bird's-eye view consisting of closed polygons of rooms and other structural elements, such as doors and windows. We trained and tested our model on the Structured3D dataset \cite{zheng2020structured3d} and showed the second-best performance compared to the state-of-the-art methods, demonstrating our model's cross-domain generalizability. 

Our main contributions are summarized as follows:
\begin{itemize}
\item We propose a novel Region-of-Interest (RoI) query-based approach for vectorized building outline extraction. We formulate each vertex as a query, which only attends to the most relevant areas and enables vertex-level interaction within the same building, thereby containing the computational burden.
\item We propose a learnable logit embedding and explore adaLN as the fusion strategy to incorporate logit and positional embeddings into the vertex queries for better vertex classification. Our approach achieves high performance in polygon quality and vertex redundancy even without post-processing.
\item By enabling efficient and effective vertex-level interaction, our proposed approach achieves excellent performance for building outlining on CrowdAI, outperforming the existing methods in most of the MS-COCO evaluation metrics, and shows superior performance on small-sized buildings. Furthermore, our approach demonstrates its generalizability with competitive performance compared to the models dedicated to 2D floorplan reconstruction on the Structured3D dataset.
\end{itemize}

\section{Related work}
\noindent\textbf{Segmentation-based methods.}
In the multi-step building outline extraction paradigm, segmentation-based methods have been widely explored to mitigate the challenges of extracting building outlines directly from aerial or satellite images.
First, semantic or instance segmentation networks are employed to generate object masks.
Then, post-processing methods are utilized to extract object polygons from these intermediate representations.
For example, \cite{zhao2018building} first uses MaskR-CNN \cite{he2017mask} to predict building segmentation masks. 
Subsequently, it utilizes the Douglas-Peucker algorithm \cite{douglas1973algorithms} to extract building polygons and applies the Minimum Descriptor Length algorithm \cite{sohn2012implicit} to regularize the building polygons iteratively. 
Recent approaches have attempted to leverage geometric features as additional supervision signals for more precise instance segmentation to facilitate the subsequent vectorization process. 
For instance, frame field learning (FFL) \cite{girard2021polygonal} learns a frame field that aligns to object tangents and improves segmentation performance, in order to facilitate the following polygonization algorithm. 
Also, HiSup \cite{xu2023hisup} predicts the attraction field map alongside the segmentation mask at the middle-level description, bridging the connection between points and regional shapes.
Despite the promising performance achieved by these multi-step methods, by design, they involve multiple models or stages, inevitably resulting in accumulated errors and operational inefficiencies during inference. 

\noindent\textbf{End-to-end methods.} 
Different from the segmentation-based methods, end-to-end methods predict building outlines directly, circumventing the step of generating intermediate representations from raw images separately. 
Building outline structures are formulated as closed polygons characterized by ordered vertex sequences.
Naturally, Recurrent Neural Networks (RNNs) are adopted to handle the various lengths of different object polygons and predict polygon vertices in an auto-regressive fashion.
Among others, Polygon-RNN \cite{castrejon2017annotating}, Polygon-RNN++ \cite{acuna2018efficient}, and PolyMapper \cite{li2019topological, zhao2021building} have shown that this sequentialization method simplifies the feature extraction step from images and enables end-to-end learning. 
Alternative to the RNN-based methods, Curve-GCN \cite{ling2019fast} proposes a Graph Convolutional Network (GCN) to simultaneously predict the locations of all the vertices, alleviating the sequential nature of polygons by a topology of a fixed number of vertices. 
PolyWorld \cite{zorzi2022polyworld} further formulates the connections between polygon vertices as a linear sum assignment problem and employs a GCN to learn the pairwise vertices connection.
However, the RNN-based methods engage the time-consuming auto-regressive process and may encounter compound errors when the initial vertices are not accurately predicted, and the bottom-up GCN-based methods rely on inefficient matrix permutations to learn connections between vertices pairwisely, even for vertices with large distances.  

\noindent\textbf{Transformers-based polygon prediction}. 
With the boom of Transformers \cite{vaswani2017attention} across domains, they are quickly adopted to polygon prediction tasks. 
Among the first transformer-based polygon prediction methods, PolyTransform \cite{liang2020polytransform} utilizes a deforming network to facilitate spatial interactions among the vertex queries of a polygon.  
However, as developed for instance segmentation, the initial polygon vertices are first sampled from the boundary of instance masks predicted by a segmentation network, and then PolyTransform predicts dense vertices for each polygon, resulting in redundant vertices. 
Polygonizer \cite{khomiakov2023polygonizer} utilizes a Transformer encoder to process image features, yet it still relies on Long Short-Term Memories (LSTMs) to predict polygon vertices, which is inefficient due to the auto-regressive process. 
Moreover, the query-based DEtection TRansformer (DETR) architecture \cite{carion2020end} is explored to treat polygon prediction as a set prediction problem for polygons.
For example, PolyBuilding \cite{hu2023polybuilding} adopts Deformable DETR \cite{zhu2020deformable} to model each polygon as a decoder query. 
It predicts a fixed number of vertices for each polygon and a classification score for each vertex. They pool out invalid vertices with scores lower than a preset threshold. 
However, it lacks vertex-level interactions in each polygon and thus requires post-processing to further remove redundant vertices, especially at building corners. 
The recent work RoomFormer \cite{yue2023connecting} employs deformable cross-attention \cite{zhu2020deformable} to allow all the vertex queries of inter and intra polygons to interact with features at reference points on the feature maps, achieving excellent prediction performance for room outline prediction.
However, this global cross-attention engages high computational costs, which is time-consuming for processing images with multiple buildings and most building polygons involve many more vertices than rooms mostly of a rectangular shape.

\section{Methodology}
\subsection{Reformulation of building polygon prediction}
Before delving into the framework of the proposed model, we reformulate building outline extraction as the problem of identifying a set of ordered vertices $V = \{v_k\}_{k=1}^G$ in each building polygon. 
G is the number of ground truth vertices, which varies in different polygons. 
To address this issue, a fixed number of $M$ vertices are uniformly sampled along the contour of a ground truth polygon, where $M$ is greater than $G$. 
To ensure that all the ground truth vertices are contained in the newly sampled vertex set $S = \{s_k\}_{k=1}^M$, PolyBuilding \cite{hu2023polybuilding} searches for a bipartite matching between the two sets that minimizes the cost $\mathcal{L}_{cost}$, as denoted by Eq.~\eqref{eq:la_polybuilding}
\begin{equation}
    \label{eq:la_polybuilding}
    \mathcal{L}_{cost} = \Sigma_{i=1}^{M}\Sigma_{j=1}^{G}C_{i,j}X_{i,j}
\end{equation}
where $C_{i,j}$ is the cost matrix. PolyBuilding formulates it as the Euclidean distance between the ground truth vertex $v_i$ and the sampled vertex $s_j$. $X_{i,j}$ is a boolean matrix, where $X_{i, j} = 1$ if $v_i$ is assigned to $s_j$. 
If $X_{i, j} = 1$, $s_j$ will be marked as valid with class 1 and shifted to the position of $v_i$ to maintain the original position of the ground truth vertices. 
Otherwise, $s_j$ will be marked as invalid with class 0. 

\begin{figure}[tb]
  \centering
  \includegraphics[trim={0 2.3cm 0 2.3cm}, clip, width=1.0\columnwidth]{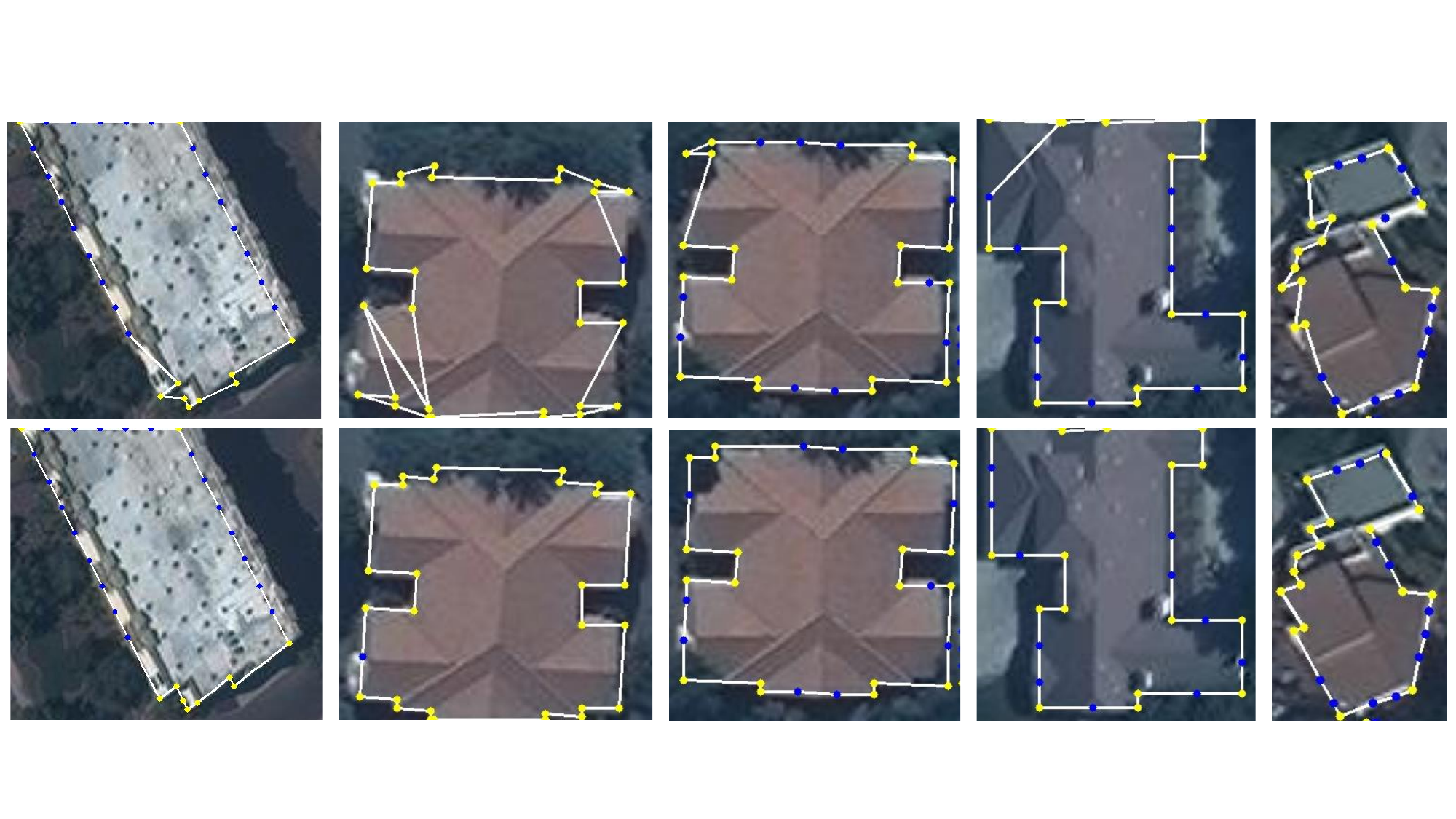}
  \caption{Distance-based matching in PolyBuilding \cite{hu2023polybuilding} (first row) vs. our ordering-based matching (second row) for vertex sampling in each polygon. The sampled vertices are marked in blue while the ground truth vertices are marked in yellow.
  }
  \label{fig:building_polygon_encoding}
  % \vspace{-10pt}
\end{figure}
However, as shown in the first row of Fig. \ref{fig:building_polygon_encoding}, we observe that this distance-based assignment causes false order permutation of the ground truth vertices, leading to edge intersections and shape distortions when the ground truth vertices are close to each other. 
To avoid this issue, we reformulate the cost matrix $C_{i,j}$ using the index difference instead of the Euclidean distance. 
Concretely, we take the top-left point on the polygon contour as the starting point and represent the contour as a dense sequence of points in counterclockwise order. 
Then, we record the indices of the $M$ sampled vertices and all the ground truth vertices in the point sequence. 
The new cost matrix between the ground truth vertex $v_i$ and the sampled vertex $s_j$ can be represented by Eq.~\eqref{eq:la_roipoly}
\begin{equation}
    \label{eq:la_roipoly}
    C_{i,j} = |p_{v_i} - p_{s_j}|
\end{equation}
where $p_{v_i}$ and $p_{s_j}$ are the indices of $v_i$ and $s_j$ in the point sequence, respectively. 
This ordering-based matching yields more accurate vertex sampling in terms of maintaining polygon shapes, as shown in the second row of Fig. \ref{fig:building_polygon_encoding}. 

%Moreover, because we have specified the ordering of a fixed number of vertices in each polygon, at the training time, our model only needs to learn the positions and class labels of the $M$ vertices in each polygon; 
%Hungarian matching is not needed as required in a standard DETR-based detection model~\cite{zhu2020deformable} for the set prediction, thus encouraging fast training.

\subsection{Architecture overview}
\begin{figure}[tb]
  \centering
  \includegraphics[trim={3.6cm 6cm 3.6cm 5.5cm}, clip, width=1.0\columnwidth]{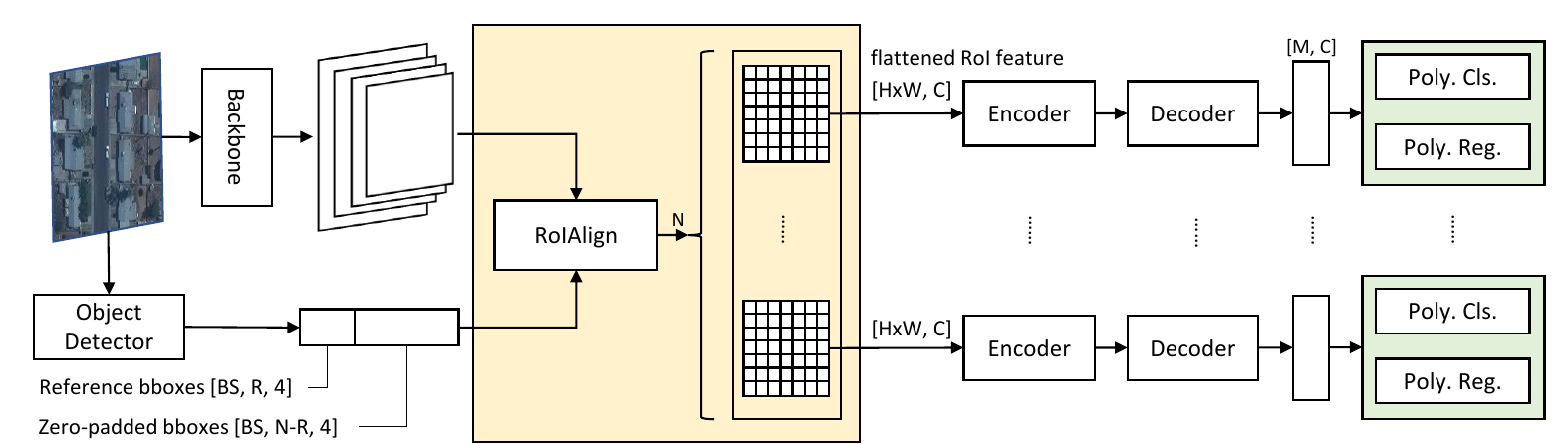}
  \caption{The overall architecture of our proposed method RoIPoly. 
  An image containing multiple buildings is fed into a backbone to extract multi-scale feature maps and an object detector to generate proposal Building Bounding Boxes (BBBs). 
  The RoIAlgin module extracts the RoI features for each building instance located by a corresponding BBB, which are later fed into an encoder-decoder pipeline with vertex and logit queries. 
  In each such pipeline, a vertex coordinate regression head and a vertex classification head are utilized to provide the final prediction simultaneously. 
  The variables denoted in this figure: $\text{BS}$ is the batch size, $R$ is the number of ground truth polygons of an image, $N$ is the number of proposal polygons per image, $M$ is the number of proposal vertices per polygon, $C$ is the feature dimension, and $H \times W$ is the RoI resolution.
  % \vspace{-6pt}
  }
  \label{fig:network_architecture}
\end{figure}

In this subsection, we introduce the overall architecture of the proposed model, named Region-of-Interest Polygon (RoIPoly).
It takes the images as input and outputs a polygon for each building.
As shown in Fig. \ref{fig:network_architecture}, RoIPoly consists of a backbone to extract multi-layer feature maps from the input image, an object detection module to generate proposal Building Bounding Boxes (BBBs) at inference time, a RoIAlign-based module to constrain the query attention on the most relevant regions of a potential building, and a query-based decoder with an additional learnable logit embedding for each polygon prediction.
It is worth mentioning that RoIPoly follows the end-to-end paradigm and does not require segmentation masks as an intermediate representation. 
We detail the RoI query attention and the learnable logit embedding in the following.

\vspace{4pt}
\noindent\textbf{Region-of-Interest (RoI) query attention.} 
\label{sec:RoI}
In each encoder-decoder pipeline, we propose constraining the query attention on the most relevant regions of each potential building located by the RoIAlign module.
The rationale behind this straightforward but effective design is that each building is characterized by an independent polygon and vertices are not connected across polygons.
Hence, unlike RoomFormer \cite{yue2023connecting}, global attention between polygon vertices are not needed.
This significantly reduces the computational complexity.

More specifically, the backbone first outputs the multi-scale feature maps $f_{l} \in \mathbb{R}^{C \times H_{l} \times W_{l}}$ extracted from the input image, where $l \in \{2, 3, 4, 5\}$ denotes the feature map levels and $f_l$ has a resolution $2^l\times$ lower than the input image, following the design of Sparse R-CNN \cite{sun2021sparse}.
To further balance the trade-off between computational cost and map resolution, we use the size $S$ of a BBB to select the most relevant feature map at layer $l^*$ by Eq.~\eqref{eq:level_assignment}, where $l^* \in \{2, 3, 4, 5\}$.
\begin{equation}
    l^* = \lfloor l_c + \log_2{\frac{S}{S_c}} \rfloor,
    \label{eq:level_assignment}
\end{equation}
where $l_c$ and $S_c$ stand for the canonical level and canonical bounding box size, respectively. 
Finally, we divide the RoI on the selected feature map $f_{l^*} \in \mathbb{R}^{C \times H_{l^*} \times W_{l^*}}$ into $H_r \times W_r$ bins. 
In this step, the RoI feature map is denoted as $r \in \mathbb{R}^{C \times H_r \times W_r}$, whose value of each bin is decided by the value of its central location through a bilinear interpolation. 
It should be noted that, alternatively, all the feature maps can be stacked to generate a more detailed RoI feature map. 
Interestingly, empirical results show that with an approximately doubled training time, the more detailed RoI does not lead to obvious performance gain (more details are presented in the ablation study \ref{sec:ablation}).

Based on the above design, we can easily calculate the computational complexity of RoIPoly in comparison with RoomFormer \cite{yue2023connecting} in terms of encoding and decoding parts.
Assuming that the number of vertex queries per polygon is $N_q$, the number of proposal polygons is $N$, the number of pixels of all level feature maps is $N_e$, the number of sampled key number is $K$ and the feature dimension is $C$, in RoomFormer \cite{yue2023connecting} the complexity of the deformable attention in the encoder is $O(2\underline{N_e}C^2+\underline{N_e}KC^2)$ and in the decoder is $O(N_q^2\underline{N^2}C^2)$.
In comparison, in RoIPoly the complexity is $O(2\underline{NH_rW_r}C^2+\underline{NH_rW_r}KC^2)$ and $O(N_q^2\underline{N}C^2)$, respectively.
In our case, the input image has a size of $300\times300$ pixels, $H_r = W_r = 7$, $N = 34$, and $N_e = 80^2 + 40^2 + 20^2 + 10^2$.
Hence, $NH_rW_r \ll N_e$ and $N \ll N^2$. 

To summarize, thanks to the constrained query attention in the RoI, we avoid the computational explosion and thus overcome the limitation of applying Transformer attention to vertex queries with vertex-level interactions in building outline extraction.

\vspace{4pt}
\noindent\textbf{Learnable logit embedding.}
To facilitate the classification head in distinguishing valid and invalid vertices in a polygon, we propose a learnable logit embedding, inspired by the positional embedding in the decoder of the DETR-based models, \textit{e.g.} \cite{carion2020end, zhu2020deformable, liu2022dab, meng2021conditional}.

\begin{figure}[t!]
\centering
    \begin{subfigure}{0.405\linewidth}
    \includegraphics[clip=true, trim=5.54cm 3.5cm 19cm 4.3cm, width=1\textwidth]{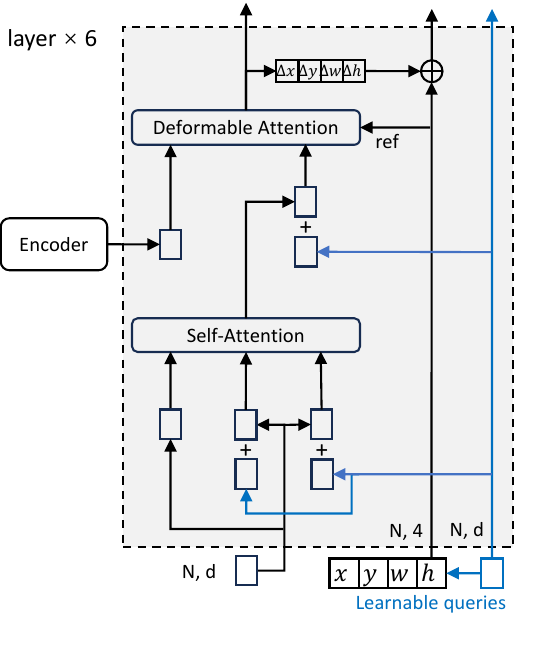}
    \caption{PolyBuilding}
    \end{subfigure}%
    \begin{subfigure}{0.59\linewidth}
    \includegraphics[clip=true, trim=14.75cm 3.5cm 5.6cm 4.3cm, width=1\textwidth]{decoder.pdf}
    \caption{RoIPoly (ours)}
    \end{subfigure}%
    \caption{Comparison of the decoder in PolyBuilding \cite{hu2023polybuilding} and RoIPoly. For clarity, we only show one layer of the decoder, omitting the sigmoid and feed-forward modules.} 
    \label{fig:decoder}
    % \vspace{-10pt}
\end{figure}

To better understand the proposed logit embedding, we briefly revisit the positional embedding in DETR.
Let $q_k$ denote the $k$-th query in the decoder and $F$ the fusion operation,
\begin{equation}
    q_k = F(e_k, p_k),
    \label{eq:q_k1}
\end{equation}
where $e_k$ and $p_k$ denote the decoder embedding and the learnable positional embedding, respectively.
In the context of vertex positional embedding, $p_k$ is mapped by
\begin{equation}
    p_k = {\rm LN}({\rm MLP}({\rm PE}({\rm sigmoid}(v_k)))),
    \label{eq:p_k1}
\end{equation}
where $v_k$ denotes the 2D coordinates of the $k$-th vertex, $\rm PE$ is the positional encoding operation to map these float coordinates into a sinusoidal embedding \cite{vaswani2017attention}, and ${\rm MLP}$ stands for a multilayer perception followed by a layer normalization $\rm LN$. 
The vertex coordinates $v_k = (x_k, y_k)$ are initialized from $\mathcal{N} (0, 1)$ followed by the sigmoid activation function, which is later used as the reference point of the corresponding decoder query $q_k$ to sample relevant attention points from the feature space in the cross-attention module. 
As denoted in Fig. \ref{fig:decoder}\,(b), this reference point is gradually refined by learning the residual $(\Delta x_k, \Delta y_k)$ layer-by-layer.
Since the outputs are real numbers, the inverse sigmoid function is applied to map the normalized coordinates back to $(-\infty, +\infty)$ and the sigmoid function is applied to map them to $[0, 1]$ again after the addition.
\begin{equation}
    v_k = {\rm sigmoid}({\rm inverse \underline{\hspace{0.5em}} sigmoid}(v_k) + {\rm MLP}^{\rm coords}(q_k)).
    \label{eq:coords_refine}
\end{equation}

In our building outline extraction task, not all the sampled vertices are valid ones (see Fig. \ref{fig:building_polygon_encoding}). 
Hence, we further incorporate the logit embedding $l_k$ as a prior into the query by extending Eq.~\eqref{eq:q_k1} to
\begin{equation}
    q_k = F(e_k, p_k, l_k),
    \label{eq:q_i}
\end{equation}
where $l_i$  is generated by
\begin{equation}
    l_i = {\rm MLP}^{\rm logits}({\rm PE}({\rm sigmoid}(\text{cls}_i))).
    \label{eq:l_i}
\end{equation}
Here, $\text{cls}_i \in (-\infty, +\infty)$ is a real number initialized from $\mathcal{N}(0, 1)$. 
We use the $\rm sigmoid$ function to map $\text{cls}_i$ to a probability value in [0, 1], indicating the confidence score of the vertex being a valid vertex.
This confidence score is used to threshold invalid vertices, which is essential to remove redundant vertices.
% Similar to the~\eqref{eq:p_k1}, sinusoidal embeddings,${\rm PE}$ denotes the sinusoidal embeddings.
Similar to the vertex, the logit $\text{cls}_i$ is refined layer-by-layer using a linear mapping in the decoder, as denoted by
\begin{equation}
    \text{cls}_i = \text{cls}_i + {\rm Linear}(q_k).
    \label{eq:cls_i}
\end{equation}
It should be noted that both $\text{cls}_i$ and ${\rm Linear}(q_i)$ are logits, hence we refer to the class embedding as logit embedding in this paper.

Considering that the vertex and logit embeddings are derived from different inputs, different from the original fusion function $F$ proposed in DETR, we design a more sophisticated function to fuse them.
We observe that the adaptive layer norm (adaLN) is widely used in GANs \cite{brock2018large, karras2019style} and diffusion models \cite{dhariwal2021diffusion} to process conditional information, which has shown better performance compared to other fusion techniques \cite{peebles2023scalable}. 
Similar to that, the learnable logit embedding serves as a conditional modulation to the attention map.
Therefore, we explore the ${\rm adaLN}$ here to incorporate it into the decoder query by regressing the scale and shift parameters $\gamma \in \mathbb{R}^C$ and $\beta \in \mathbb{R}^C$ from the learnable logit embedding. 
The fusion process is formulated as
\begin{equation} \label{eq1}
\begin{split}
\gamma, \beta &= {\rm adaLN}(l_k), \\
    q_k &= (e_k + p_k) \times (\gamma + 1) + \beta.
\end{split}
\end{equation}

Fig. \ref{fig:decoder} shows the comparison between PolyBuilding \cite{hu2023polybuilding} with a standard DETR decoder and our RoIPoly with the additional learnable logit embedding. We refer the readers to \cite{carion2020end,zhu2020deformable} for more detailed information about the self- and cross-attention in the DETR decoder. 

\subsection{Loss functions}
During training, the number of ground truth building instances $N^{\rm gt}$ per image is known, where $N^{\rm gt} \leq N$ and varies for different images. 
Hence, for each image, we only leverage the first $N^{\rm gt}$ pairs of predicted and reference polygons for calculating the vertex coordinate regression loss, which is a standard $\mathcal{L}_1$ loss defined as
\begin{equation}
\label{loss:cor}
    \mathcal{L}_{\rm cor} = \frac{1}{N^{\rm gt}\times M} \sum_{n=1}^{N^{\rm gt}} d(V_n, V_n^*),
\end{equation}
where $V_n = \{ v_{n, 1}, v_{n, 2}, ..., v_{n, M} \}$ is the padded ground truth vertex sequence and $V_n^* = \{ v^*_{n, 1}, v^*_{n, 2}, ..., v^*_{n, M} \}$ is the predicted vertex sequence of the $n\text{-}th$ polygon.

We use the focal loss \cite{lin2017focal} to calculate the classification loss of all the $N$ pairs of predicted and ground truth polygons.
\begin{equation}
\label{loss:cls}
    \mathcal{L}_{\rm cls} = \frac{1}{N \times M} \sum_{n=1}^{N}\sum_{m=1}^{M} {\rm Focal}(\lambda_{m,n}^*, \lambda_{m,n}),
\end{equation}
where $\lambda_{m,n}^*$ and $\lambda_{m,n}$ are the predicted and the corresponding ground truth vertex class labels, respectively.

Finally, the total loss is the weighted summation of $\mathcal{L}_{\rm cor}$ and $\mathcal{L}_{\rm cls}$, defined as
\begin{equation}
    \mathcal{L}= \lambda_{\rm cor}\mathcal{L}_{\rm cor} + \lambda_{\rm cls}\mathcal{L}_{\rm cls},
\end{equation}
where $\lambda_{\rm cor}$ and $\lambda_{\rm cls}$ are the corresponding coefficients.

\section{Experiment}
\subsection{Experimental setup}
\noindent\textbf{Dataset.} 
We evaluated RoIPoly on the mainstream dataset CrowdAI Mapping Challenge dataset for vectorized building outline extraction \cite{mohanty2020deep}. 
It contains 280,741 and 60,317 satellite images of a size of $300\times300$ pixels for training and testing, respectively. 
Polygon annotations are in MS-COCO format \cite{lin2014microsoft}. 

We also trained and tested our model on the popular Structured3D dataset \cite{zheng2020structured3d} for 2D floorplan reconstruction, which provides photo-realistic images covering 3,500 diverse indoor scenes with semantically rich structure annotations. It contains 3,000 training samples, 250 validation samples, and 250 test samples. We adhere to the existing methods \cite{yue2023connecting, chen2022heat, stekovic2021montefloor} to process the dataset. To be specific, we first convert the registered multi-view RGB-D panoramas to 3D point clouds. Then we project them along the vertical axis into a bird's-eye view 2D density map of size 256$\times$256 pixels, where the value of each pixel on the density map is the normalized number of projected points to the pixel. 

\noindent\textbf{Evaluation metrics.} 
We evaluate the generated building polygons considering three aspects: building instance segmentation, polygon quality, and vertex redundancy. 
Building instance segmentation is measured by Intersection over Union (IoU) and MS-COCO metrics \cite{lin2014microsoft}, including Average Precision (AP) and Average Recall (AR) based on different IoU thresholds, which are also the official evaluation criteria of the CrowdAI dataset.
Fidelity and regularity is measured in ${\rm AP_{boundary}}$ \cite{cheng2021boundary}, PoLiS \cite{avbelj2014metric}, and Max Tangent Angle Error (MTA) \cite{girard2021polygonal}. 
${\rm AP_{boundary}}$ is calculated based on the boundary IoU that measures the boundary overlap of two masks.
PoliS measures the shape differences by calculating the shortest distance between two polygons. 
MTA evaluates the geometric regularity by calculating the tangent angle difference between predicted and ground truth polygons.
Vertex redundancy is measured in N ratio \cite{zorzi2022polyworld} and complexity aware IoU (C-IoU) \cite{zorzi2022polyworld}. 
N ratio is the ratio between the vertex number of the predicted and ground truth polygons -- the closer to 1, the better the N ratio is. 
A higher C-IoU indicates a better trade-off between segmentation accuracy and vertex redundancy.

On the Structured3D dataset, we leverage the same evaluation metrics as existing methods  \cite{yue2023connecting, chen2022heat, stekovic2021montefloor}. We first find the best-matching rooms based on the IoU between the ground truth and the predictions and then compute precision, recall, and F1 scores for rooms, corners, and angles. 

\noindent\textbf{Implementation Details.}
On the CrowdAI dataset, we use the Feature Pyramid Network (FPN) \cite{lin2017feature} to generate multi-scale feature maps for ResNet50 \cite{he2016deep} and Swin \cite{liu2021swin} backbones following \cite{chen2023diffusiondet}. 
All feature maps have $C = 256$ channels. 
The encoder and decoder contain six layers with 256 channels each. 
We use eight heads in the self-attention and the deformable cross-attention module, and four reference points for each query in the deformable cross-attention module. 
To accelerate training and reduce computational burden, we handle small, medium (${\rm area}< 96^2$ pixels) and large buildings (${\rm area} \geq 96^2$ pixels) separately. 
For small and medium buildings, the number of proposal vertices per polygon and proposal polygons per image is set to $M = 30$ and $N = 34$. 
While, for large buildings, $M = 96$ and $N = 10$. 
On the Structured3D dataset, we only use ResNet-50 as the backbone and align the training settings with RoomFormer \cite{yue2023connecting} for a fair comparison. 

\begin{table}[tb]
  \caption{Comparison with state-of-the-art models on CrowdAI in MS-COCO evaluation metrics. Best values are in boldface and second-best values are underlined.
  }
  \label{tab:MS_COCO}
  \centering
  \resizebox{\linewidth}{!}{
  % \setlength{\tabcolsep}{1pt}
  % \fontsize{8pt}{8pt}\selectfont
  \begin{tabular}{@{}l|c|cccccccccccc@{}}
    \toprule
    Method & Backbone & $\rm AP \uparrow$ & $\rm AP_{50} \uparrow$ & $\rm AP_{75} \uparrow$ & $\rm AP_{S} \uparrow$ & $\rm AP_{M} \uparrow$ & $\rm AP_{L} \uparrow$ & $\rm AR \uparrow$ & $\rm AR_{50} \uparrow$ & $\rm AR_{75} \uparrow$ & $\rm AR_{S} \uparrow$ & $\rm AR_{M} \uparrow$ & $\rm AR_{L} \uparrow$ \\
    \midrule
    PolyMapper \cite{li2019topological} & VGG16 & 55.7 & 86.0 & 65.1 & 30.7 & 68.5 & 58.4 & 62.1 & 88.6 & 71.4 & 39.4 & 75.6 & 75.4 \\
    FFL \cite{girard2021polygonal} & UResNet101 & 61.3 & 87.5 & 70.6 & 33.9 & 75.1 & 83.1 & 64.9 & 89.4 & 73.9 & 41.2 & 78.7 & 85.9 \\
    PolyWorld \cite{zorzi2022polyworld} & R2U-Net & 63.3 & 88.6 & 70.5 & 37.2 & 83.6 & \underline{87.7} & 75.4 & 93.5 & 83.1 & 52.5 & 88.7 & \underline{95.2} \\
    PolyBuilding \cite{hu2023polybuilding} & ResNet50 & 78.7 & \textbf{96.3} & \underline{89.2} & \text{-} & \text{-} & \text{-} & 84.2 & 97.3 & 92.9 & \text{-} & \text{-} & \text{-} \\
    HiSup \cite{xu2023hisup} & HRNetV2-W48 & 79.4 & 92.7 & 85.3 & 54.3 & \textbf{91.6} & \textbf{96.5} & 81.5 & 93.1 & 86.7 & 59.3 & \textbf{93.7} & \textbf{97.7} \\ \midrule
    RoIPoly (ours) & ResNet50 & \underline{81.4} & 95.8 & 89.1 & \underline{72.0} & 87.5 & 80.3 & \underline{87.8} & \underline{97.7} & \underline{94.0} & \underline{81.0} & 91.6 & {91.4} \\
    RoIPoly (ours) & Swin-Base & \textbf{83.6} & \underline{96.0} & \textbf{90.9} & \textbf{74.2} & \underline{89.6} & 82.1 & \textbf{89.3} & \textbf{97.8} & \textbf{94.6} & \textbf{82.6} & \underline{93.0} & 92.2 \\
  \bottomrule
  \end{tabular}
  }
  % \vspace{-4pt}
  
\end{table}
\begin{table}[tb]
  \caption{Comparison with state-of-the-art models on CrowdAI in polygon quality and vertex redundancy. Best values are in boldface and second-best values are underlined.
  }
  \label{tab:other_metrics}
  \centering
  % \resizebox{\linewidth}{!}{
  \fontsize{5.6pt}{5.6pt}\selectfont
  \begin{tabular}{@{}l|c|c|ccccccc@{}}
    \toprule
    Method & Backbone & post-proc. & $\rm AP_{boundary}$$\uparrow$ & $\rm IoU$$\uparrow$ & $\rm C\text{-}IoU$$\uparrow$ & $\rm N \,ratio$$\rightarrow$1 & $\rm MTA$$\downarrow$ & $\rm PoLiS$$\downarrow$ \\
    \midrule
    PolyBuilding \cite{hu2023polybuilding} & ResNet50 & \text{NMS} & \text{-} & \underline{94.0} & \underline{88.6} & \textbf{+0.01} & \underline{32.4} & \text{-} \\
    PolyWorld \cite{zorzi2022polyworld} & R2U-Net & \text{NMS} & 50.0 & 91.2 & 88.3 & -0.07 & 32.9 & 0.96 \\
    HiSup \cite{xu2023hisup} & HRNetV2-W48 & \text{poly. simp.} & \underline{66.5} & \textbf{94.3} & \textbf{89.6} & \underline{+0.02}  & \underline{32.4} & \textbf{0.73} \\ \midrule
    
    RoIPoly (ours) & ResNet50 & \text{none} & \textbf{66.7} & 92.2 & 86.3 & +0.06 & \underline{28.6} & 0.90 \\
    RoIPoly (ours) & Swin-Base & \text{none} & 66.1 & 92.9 & 87.6 & +0.04 & \textbf{27.6} & \underline{0.79} \\
  \bottomrule
  \end{tabular}
  % }
  % \vspace{-8pt}
\end{table}

\subsection{Experimental results and analysis}
\noindent\textbf{Results on CrowdAI dataset.}
We report the performances of building outline extraction on the CrowdAI dataset using MS-COCO evaluation metrics in Table \ref{tab:MS_COCO}. 
It shows that our model RoIPoly with the RetNet50 backbone 
outperforms the end-to-end models PolyMapper \cite{li2019topological}, PolyWorld \cite{zorzi2022polyworld}, and DETR-based model PolyBuilding \cite{hu2023polybuilding} in most of the metrics, except for $\rm AP_{L}$ and $\rm AR_{L}$.
It also notably outperforms the segmentation-based models FFL \cite{girard2021polygonal} and HiSup \cite{xu2023hisup} measured in the overall $\rm AP$ and $\rm AR$.
Moreover, after switching to a more powerful backbone, Swin-Base, RoIPoly's performance has improved further. 
It achieves the best performance in most of the MS-COCO evaluation metrics, improving both the $\rm AP$ and $\rm AR$ by over 4\%.
It is worth mentioning that RoIPoly achieves superior performance on small-sized buildings with the $\rm AP$ and $\rm AR$ exceeding 70\% and 80\%, respectively, while all the other models struggle to reach 60\%.
This indicates the excellent ability of RoIPoly with the RoI query to accurately locate the vertices of small buildings. 
However, we notice that HiSup significantly outperforms all the other models, including RoIPoly in $\rm AP_{L}$ and $\rm AR_{L}$.
This is because HiSup leverages instance segmentation masks for the polygonization and further applies polygon simplification with edge merging, which is more effective on large buildings with more detailed edge information.

Table \ref{tab:other_metrics} shows the performances in terms of polygon quality and vertex redundancy measured in $\rm IoU$, $\rm AP_{boundary}$, $\rm C\text{-}IoU$, $\rm N\text{-}ratio$, $\rm MTA$, and $\rm PoLiS$. 
Even compared to the models with post-processing, such as non-maximum suppression (NMS) \cite{neubeck2006efficient} and polygon simplification (poly. simp.), RoIPoly with ResNet50 achieves evidently lower $\rm MTA$, indicating a better performance of geometric regularity than the other models.
It also demonstrates a slightly better performance of $\rm AP_{boundary}$ measuring the overlap between the predicted and corresponding ground truth polygons at boundaries.
However, RoIPoly falls behind HiSup in the $\rm IoU$ related metrics and $\rm PoLiS$.  
This decline in performance stems from the comparatively inferior results tested on large buildings. 
A more detailed discussion can be found in Sec.~\ref{sec:limitations}. 
%Such buildings necessitate more queries in the RoIPoly model, but small- and medium-sized buildings dominate the CrowdAI dataset, thus amplifying training complexity,
%whereas large buildings are associated with large masks with more detailed information in HiSup for the polygonization.

\begin{figure}[t!]
\begin{subfigure}{\linewidth}
\centering
\includegraphics[clip=true, trim=0.2cm 6cm 0.2cm 6cm, width=\linewidth]{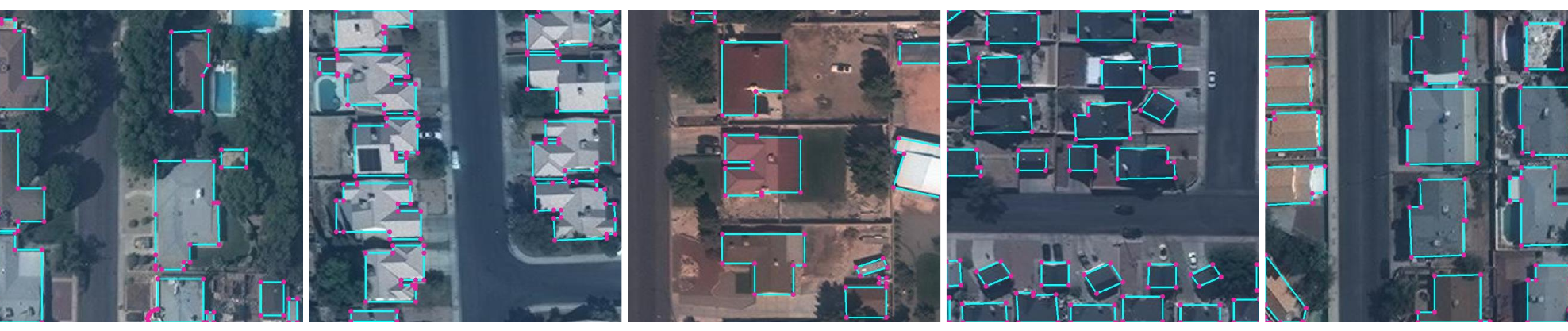}
\caption{Ground truth}
\end{subfigure}

\begin{subfigure}{\linewidth}
\centering
\includegraphics[clip=true, trim=0.2cm 6cm 0.2cm 6cm, width=\linewidth]{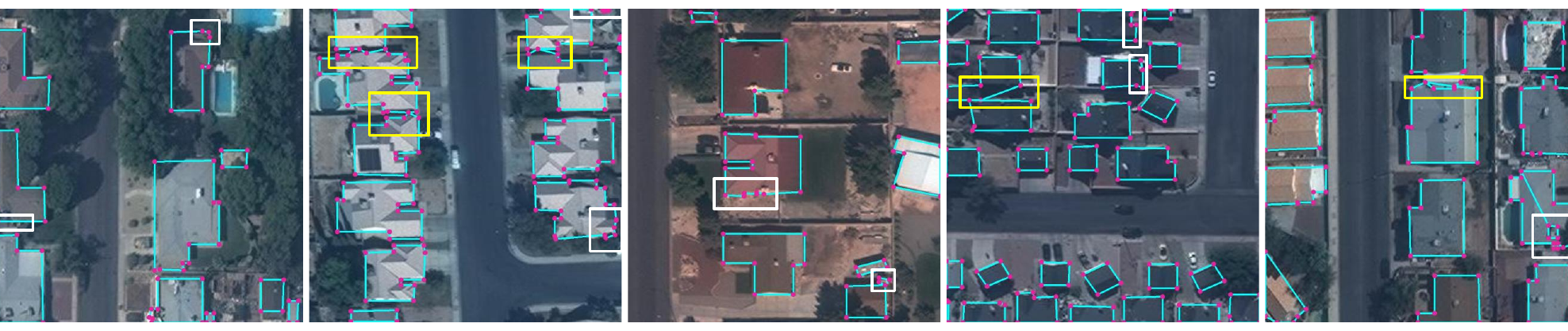}
\caption{HiSup}
\end{subfigure}

\begin{subfigure}{\linewidth}
\centering
\includegraphics[clip=true, trim=0.2cm 6cm 0.2cm 6cm, width=\linewidth]{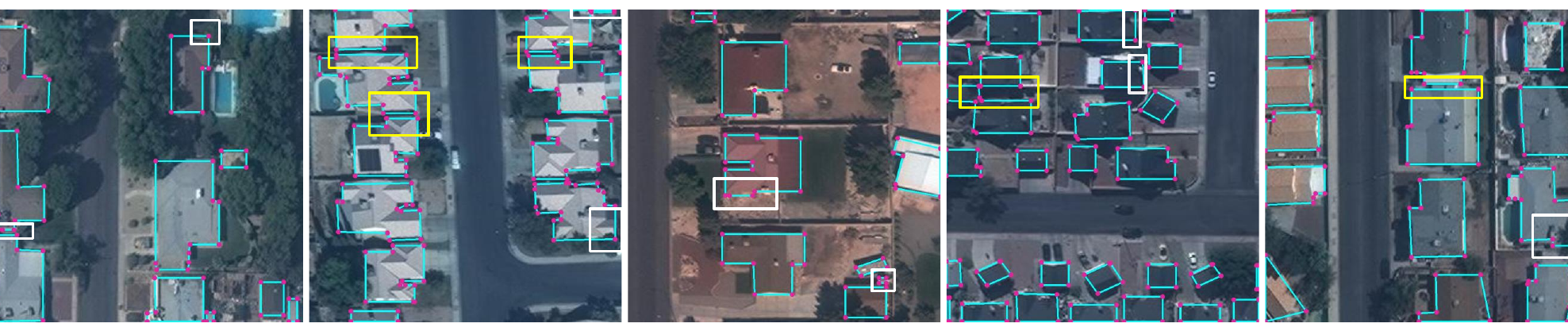}
\caption{RoIPoly (ours)}
\end{subfigure}

\caption{Examples of vectorized building polygon extraction on the CrowdAI dataset. Top row: Ground truth. Middle row: HiSup \cite{xu2023hisup}. Bottom row: RoIPoly (ours). The white boxes highlight the areas that include intricate artifacts of edges and redundant vertices, and the yellow boxes highlight the areas of mixing vertices from different buildings.}
\label{fig:viz}
\vspace{-4pt}
\end{figure}

\begin{figure}[tb] 
\begin{subfigure}{0.32\linewidth}
\centering
\includegraphics[clip=true, trim=0.cm 12.3cm 22.6cm 1.0cm, width=\linewidth]{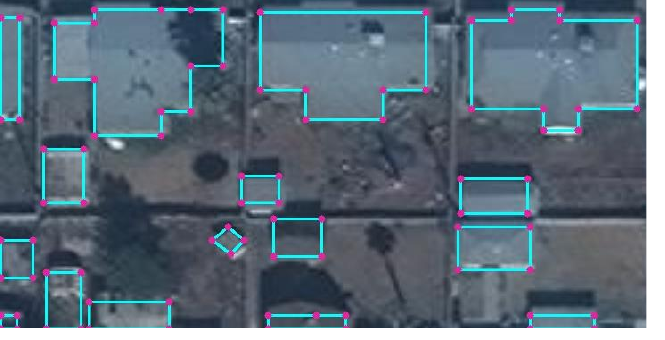}
\caption{Ground truth}
\end{subfigure}
\begin{subfigure}{0.32\linewidth}
\centering
\includegraphics[clip=true, trim=11.3cm 12.3cm 11.3cm 1.0cm, width=\linewidth]{small_buildings.pdf}
\caption{HiSup}
\end{subfigure}
\begin{subfigure}{0.32\linewidth}
\centering
\includegraphics[clip=true, trim=22.6cm 12.3cm 0cm 1.0cm, width=\linewidth]{small_buildings.pdf}
\caption{RoIPoly}
\end{subfigure}  
  \caption{Comparison of the predicted polygons of small- and medium-sized buildings. 
  % \vspace{-2pt}
  }
  \label{fig:small_buildings}
\end{figure}
Fig. \ref{fig:viz} illustrates the examples of qualitative results on the CrowdAI dataset. 
We compare the building polygons generated by RoIPoly with those by HiSup, the current state-of-the-art model tested on the CrowdAI dataset. 
It can be seen that, in general, RoIPoly produces cleaner and more regular polygons, which correctly contour the buildings appearing in the satellite imagery and are well aligned with the ground truth polygons. 
In comparison, as highlighted in color-coded boxes, HiSup is prone to generate intricate artifacts of edges and redundant vertices at corners (white boxes).
This may be due to the feature confusion induced by the additional geometric supervision signals utilized in HiSup.
Moreover, it can be often noted that HiSup mixes some vertices of different buildings that are closely located to each other (yellow boxes).
On the other hand, thanks to the RoI-constrained query attention that only allows vertex-level interactions within the same polygon, RoIPoly can easily distinguish buildings that are close to each other and avoid such issues.

We further show the comparison of the predicted polygons of small-sized buildings between HiSup and RoIPoly in Fig. \ref{fig:small_buildings}. As highlighted in the yellow boxes, RoIPoly can accurately extract the polygons for clustered small-sized buildings owing to the query attention of learning the reference points. Besides, RoIPoly leverages the high-resolution feature map, which benefits the polygon extraction of small buildings due to enriched information.
Whereas, HiSup highly relies on the quality of segmentation masks, which may result in poor performance by a slight distortion or shift of the masks on small-sized buildings. 

\noindent\textbf{Results on Structured3D dataset.} 
As shown in Table \ref{tab:stru3d}, compared to the 
state-of-the-art methods dedicated to floorplan prediction, RoIPoly achieves the second-best performance in most of the evaluation metrics, outperforming HAWP \cite{xue2020holistically}, LEAR [citation], and HEAT \cite{chen2022heat}.
% RoIPoly achieves comparable performance with the state-of-the-art methods, including HAWP \cite{xue2020holistically}, HEAT \cite{chen2022heat} and RoomFormer \cite{yue2023connecting}. 
However, RoIPoly slightly lags behind RoomFormer \cite{yue2023connecting}. 
We provide a further discussion on this performance difference in Sec.~\ref{sec:limitations}. 
%Our conjuncture is that the different layout characteristics of buildings and rooms lead to RoIPoly's performance difference in these two different tasks. 
%Unlike buildings in the CrowdAI dataset, rooms in a 2D floorplan are adjacent to each other, and the bounding box of a single room may encompass parts from multiple rooms. 
%Consequently, the attention space inside the region of interest of a single room includes irrelevant points from neighboring rooms, which causes confusion.
\begin{table}[tb]
  \caption{Comparison with the state-of-the-art models on Stuctured3d. Best values are in boldface, and second-best values are underlined.}
  \label{tab:stru3d}
  \centering
  % \resizebox{\linewidth}{!}{
  \fontsize{7pt}{6pt}\selectfont
  \begin{tabular}{@{}c|ccc|ccc|ccc@{}}
    \toprule
    \multicolumn{1}{c|}{} & \multicolumn{3}{c|}{Room} & \multicolumn{3}{c|}{Corner} & \multicolumn{3}{c}{Angle} \\
    Method & Prec. & Rec. & F1 & Prec. & Rec. & F1 & Prec. & Rec. & F1 \\
    \midrule
     HAWP \cite{xue2020holistically} & 77.7 & 87.6 & 82.3 & 65.8 & 77.0 & 70.9 & 59.9 & 69.7 & 64.4 \\
     LETR \cite{xu2021line} & 94.5 & 90.0 & 92.2 & 79.7 & 78.2 & 78.9 & 72.5 & 71.3 & 71.9 \\
     HEAT \cite{chen2022heat} & \underline{96.9} & 94.0 & 95.4 & 81.7 & \underline{83.2} & 82.5 & 77.6 & \underline{79.0} & \underline{78.3} \\
     RoomFormer \cite{yue2023connecting} & \textbf{97.9} & \textbf{96.7} & \textbf{97.3} & \textbf{89.1} & \textbf{85.3} & \textbf{87.2} & \textbf{83.0} & \textbf{79.5} & \textbf{81.2} \\
     \midrule
     RoIPoly (Ours) & 96.0 & \underline{94.8} & \underline{95.5} & \underline{84.6} & 82.7 & \underline{83.6} & \underline{79.0} & 77.3 & \underline{78.3} \\
  \bottomrule
  \end{tabular}
  % }
  %\vspace{-8pt}
\end{table}

\begin{figure}[t!]
\begin{subfigure}{\linewidth}
\centering
\includegraphics[clip=true, trim=0.2cm 6cm 0.2cm 6cm, width=\linewidth]{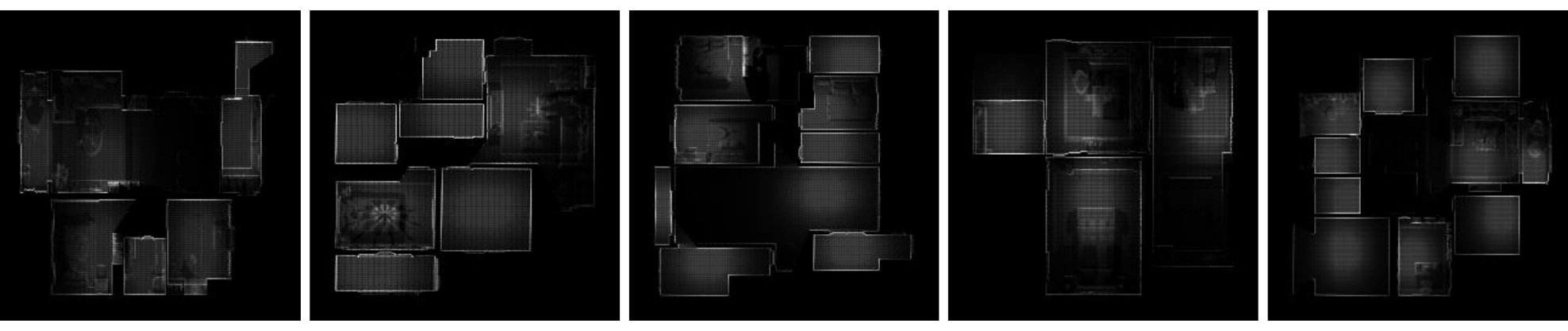}
\caption{Density map}
\end{subfigure}

\begin{subfigure}{\linewidth}
\centering
\includegraphics[clip=true, trim=0.2cm 6cm 0.2cm 6cm, width=\linewidth]{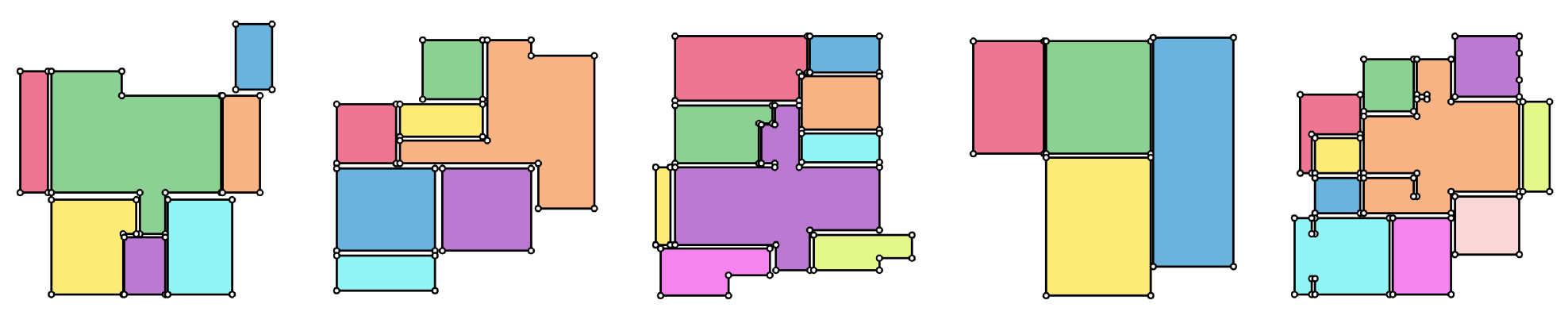}
\caption{Ground truth}
\end{subfigure}

\begin{subfigure}{\linewidth}
\centering
\includegraphics[clip=true, trim=0.2cm 6cm 0.2cm 6cm, width=\linewidth]{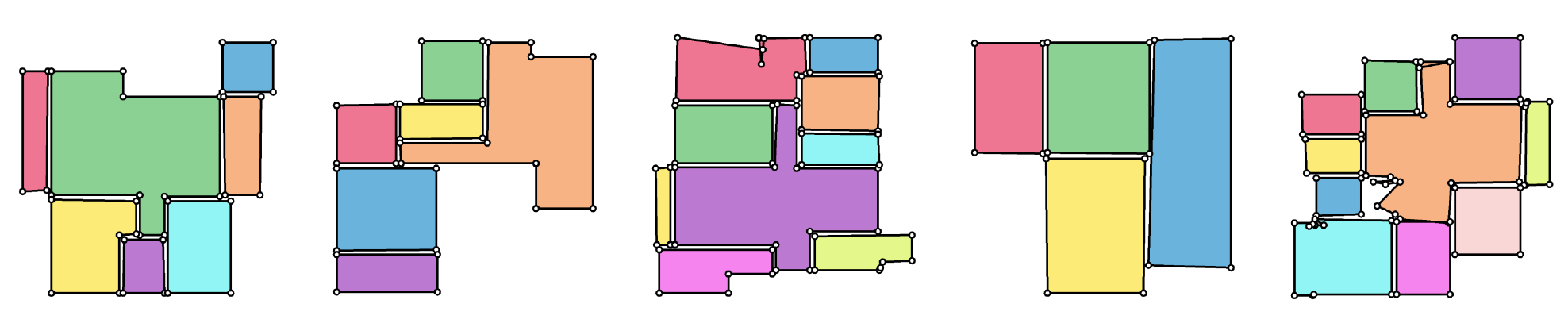}
\caption{RoomFormer}
\end{subfigure}

\begin{subfigure}{\linewidth}
\centering
\includegraphics[clip=true, trim=0.2cm 6cm 0.2cm 6cm, width=\linewidth]{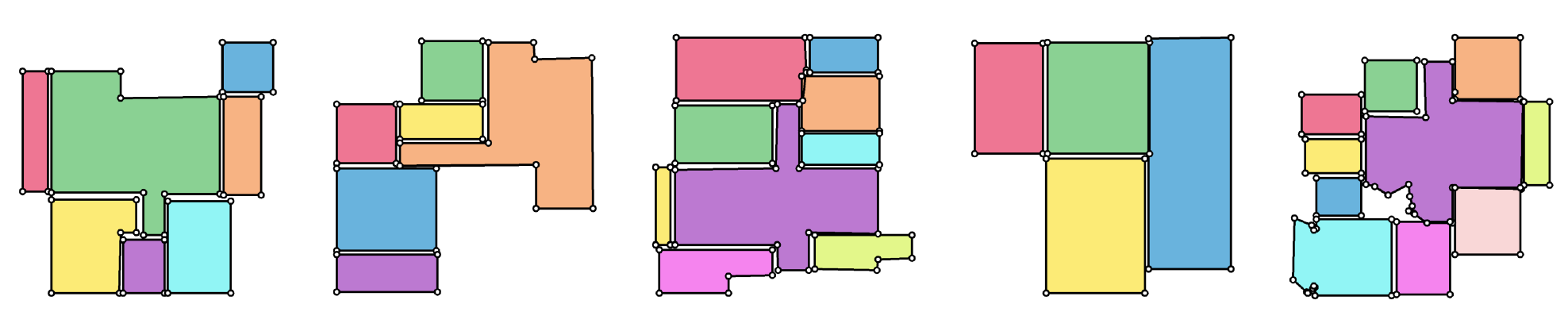}
\caption{RoIPoly (ours)}
\end{subfigure}
\caption{Examples of reconstructed polygonal floorplans on the Structured3D dataset \cite{zheng2020structured3d}. First row: 2D density map of the 3D point cloud data. Second row: Ground truth. Third row: RoomFormer \cite{yue2023connecting}. Fourth row: RoIPoly (ours).}
\label{fig:stru3d}
\end{figure}

Further qualitative results are shown in Fig. \ref{fig:stru3d}. 
As can be seen, both RoIPoly and RoomFormer perform well on regular room types. 
However, both methods tend to produce intricate artifacts when encountering rooms with narrow and concave polygonal shapes. This could be attributed to the close proximity of the vertices on these edges, making it difficult to effectively distinguish them in the feature map.

\subsection{Ablation study}
\label{sec:ablation}
% \vspace{4pt}
% \noindent\textbf{Learnable logit embedding.}
The ablation study mainly focuses on validating the effectiveness of the learnable logit embedding, the inclusion of multi-scale feature maps into the RoI, and the effectiveness of adaLN. 
For simplicity, we conduct the comparison experiments on the small and medium-sized buildings, which account for over 90\% of all the buildings. 
All the above models were trained for 60 epochs without any learning rate drop. 
We also compare the computational complexity between using global attention in RoomFormer \cite{yue2023connecting} and constraining attention within the RoI in RoIPoly. The input resolution is set to 320$\times$320. 

\begin{table}[tb]
  \caption{The effectiveness of the proposed learnable logit embedding.}
  \label{tab:ll}
  \centering
  \resizebox{\linewidth}{!}{
  % \fontsize{6pt}{6pt}\selectfont
  \begin{tabular}{@{}cc|ccccccccc@{}}
    \toprule
    Logit emb. & $\rm GFLOPs\,(G)$$\downarrow$ & $\rm Params\,(M)$$\downarrow$ & $\rm AP$$\uparrow$ & $\rm AP_{50}$$\uparrow$ & $\rm AP_{75}$$\uparrow$ & $\rm IoU$$\uparrow$ & $\rm C\text{-}IoU$$\uparrow$ & $\rm N \,ratio$$\rightarrow$1 & $\rm MTA$$\downarrow$ & $\rm PoLiS$$\downarrow$ \\
    \midrule
    $\text{-}$  & 14.9 & 9.8 & 72.6 & 90.3 & 80.3 & 89.5 & 84.2 & -0.08 & 30.0 & 0.92 \\
    %$\checkmark$ & 29.3 & 14.3 & 73.9 & 90.7 & 82.0 & 89.9 & 84.9 & -0.07 & 29.9 & 0.90  \\
    $\checkmark$ & 29.3 & 14.3 & {77.5} & {94.2} & {86.7} & {90.8} & {86.5} & {+0.01} & {29.6} & {0.87} \\
  \bottomrule
  \end{tabular}}
  \vspace{-8pt}
\end{table}

\begin{table}[tb]
  \caption{Comparison between using RoI from one-level and all-level feature maps.}
  \label{tab:alllevels}
  \centering
  \resizebox{\linewidth}{!}{
  \fontsize{6pt}{6pt}\selectfont
  \begin{tabular}{@{}c|cc|cccccccc@{}}
    \toprule
    All level RoI & $\rm GFLOPs\,(G)$$\downarrow$ & $\rm Params\,(M)$$\downarrow$ & $\rm AP$$\uparrow$ & $\rm AP_{50}$$\uparrow$ & $\rm AP_{75}$$\uparrow$ & $\rm IoU$$\uparrow$ & $\rm C\text{-}IoU$$\uparrow$ & $\rm N \,ratio$$\rightarrow$1 & $\rm MTA$$\downarrow$ & $\rm PoLiS$$\downarrow$ \\
    \midrule
    $\text{-}$  & 29.3 & 14.3 & 77.5 & 94.2 & 86.7 & 90.8 & 86.5 & +0.01 & 29.6 & 0.87 \\
    $\checkmark$ & 55.1 & 15.2 & 77.6 & 94.0 & 85.7 & 90.8 & 86.7 & +0.02 & 29.6 & 0.85 \\
  \bottomrule
  \end{tabular}}
  % \vspace{-2pt}
\end{table}

%\begin{table}[tb]
%  \caption{Performance differences of different confidence score thresholds to remove invalid vertices.}
%  \label{tab:confthreshold}
%  \centering
  % \resizebox{\linewidth}{!}{
%  \fontsize{6pt}{6pt}\selectfont
%  \begin{tabular}{@{}c|cccccccc@{}}
%    \toprule
%    Conf. score thresh. & $\rm AP$$\uparrow$ & $\rm AP_{50}$$\uparrow$ & $\rm AP_{75}$$\uparrow$ & $\rm IoU$$\uparrow$ & $\rm C\text{-}IoU$$\uparrow$ & $\rm N \,ratio$$\rightarrow$1 & $\rm MTA$$\downarrow$ & $\rm PoLiS$$\downarrow$ \\
%    \midrule
%    $0.2$  & 82.7 & 98.1 & 93 & 92.1 & 68.3 & +0.76 & 30.0 & 0.80 \\
%    $0.3$ & 82.2 & 97.0 & 91.8 & 92.0 & 82.5 & +0.24 & 29.8 & 0.81 \\
%     $\underline{0.4}$ & 77.6 & 94.0 & 85.7 & 90.8 & 86.7 & +0.02 & 29.6 & 0.85 \\
%    $0.5$ & 53.8 & 77.0 & 57.3 & 81.0 & 70.6 & -0.22 & 31.6 & 1.26 \\
%  \bottomrule
%  \end{tabular}
  % }
%  \vspace{-8pt}
%\end{table}

As shown in the first row of Table \ref{tab:ll}, after removing the learnable logit embedding, the model's performance drops significantly in all the metrics, indicating its great importance for the improvement of vertex regression and classification performance. 

We also evaluate the effectiveness of the RoI feature map from all-level feature maps. 
Compared to what is proposed in Sec.~\ref{sec:RoI}, we stack the multi-scale feature maps to generate an RoI feature map for each building instance. As shown in Table \ref{tab:alllevels}, by considering all-level feature maps, the $\rm GFLOPs$ is almost doubled from 29.3\,G to 55.1\,G.
However, the overall performance is even slightly decreased.
Hence, we only use the most relevant feature map selected following Eq.~\eqref{eq:level_assignment}.

%Furthermore, we compare the performances of using different confidence score thresholds to remove invalid vertices in Table \ref{tab:confthreshold}. 
%It can be seen that there is a trade-off between AP and $\rm C-IoU$ and $\rm N\,ratio$. The latter two metrics mainly reflect the quality of vertex redundancy.
%Hence, we take 0.4 as the final confidence score threshold to remove invalid vertices.

Furthermore, we analyze the effectiveness of the adaLN module utilized in RoIPoly. Considering that the learnable positional embedding is simply added to the decoder query to provide positional priors in DETR-based models \cite{carion2020end, zhu2020deformable}, we replace adaLN by addition in this experiment. 
As can be seen in Table \ref{tab:adaln}, the replacement leads to an overall performance drop, which proves the effectiveness of adaLN. 
\begin{table}[tb]
  \caption{Comparison between using adaLN and addition in fusing learnable logit embeddings and other embeddings.}
  \label{tab:adaln}
  \centering
  % \resizebox{\linewidth}{!}{
  \fontsize{7pt}{6pt}\selectfont
  \begin{tabular}{@{}c|cccccccc@{}}
    \toprule
    fusing tech. & $\rm AP$$\uparrow$ & $\rm AP_{50}$$\uparrow$ & $\rm AP_{75}$$\uparrow$ & $\rm IoU$$\uparrow$ & $\rm C\text{-}IoU$$\uparrow$ & $\rm N \,ratio$$\rightarrow$1 & $\rm MTA$$\downarrow$ & $\rm PoLiS$$\downarrow$ \\
    \midrule
    addition  & 75.1 & 92.4 & 83.5 & 81.0 & 70.6 & -0.22 & 30.1 & 0.90 \\
     adaLN & 77.5 & 94.2 & 86.7 & 90.8 & 86.5 & +0.01 & 29.6 & 0.87 \\
  \bottomrule
  \end{tabular}
  % }
  % \vspace{-8pt}
\end{table}

\begin{table}[ptb]
  \caption{Comparison of the model complexity between using global attention and constraining attention within the RoI.}
  \label{tab:gflops}
  \centering
   \fontsize{7pt}{6pt}\selectfont
  \begin{tabular}{@{}c|cc@{}}
    \toprule
    Input & $\rm FLOPs\,(G)$$\downarrow$ & $\rm Params\,(M)$$\downarrow$ \\
    \midrule
    Global attention & 64.2 & 15.2 \\
    Constrained query attention (ours) & 29.3 & 14.3 \\
  \bottomrule
  \end{tabular}
  \vspace{-8pt}
\end{table}

Lastly, we compare the model complexity between using global attention in RoomFormer and our constrained query attention. 
RoomFormer adopts the entire feature map for query attention and enables global interaction among vertex queries, which leads to computational explosion when applied to building outline extraction, as the number of vertices grows drastically compared to 2D floorplan reconstruction. 
As seen in Table \ref{tab:gflops}, constraining attention within the RoI reduces the $\rm FLOPs$ by more than half, and the model size is also smaller than that of using global attention. 

\section{Discussion}
\label{sec:limitations}
Despite the great performance on the CrowdAI and the Structured3d dataset, RoIPoly still exhibits some limitations. Firstly, RoIPoly tends to generate inaccurate outlines for large buildings with complex edges. 
One potential reason is the inadequate amount of large building samples in the CrowdAI dataset, where small and medium buildings account for more than 90\%. 
Besides, we use a RoI feature of size 7$\times$7 for buildings of different sizes. While this size may be sufficient for small and medium-sized buildings, it might be inadequate for large buildings to capture sufficient geometrical information. 
In comparison, large buildings are associated with large masks with more detailed information in HiSup for the polygonization. 
To address this issue, we want to explore enhancing the resolution of the RoI and leveraging geometrical features as additional supervision signals in future work. 
%Another limitation is the dependency on the object detection module to provide proposal bounding boxes. 
%However, the issue can be easily addressed by utilizing additional prediction heads to predict bounding boxes and polygons simultaneously. 

Secondly, RoIPoly shows slightly inferior performance in 2D floorplan reconstruction compared to RoomFormer. 
We conjecture that the different layout characteristics of buildings and rooms lead to RoIPoly's performance difference in these two different tasks. 
Unlike buildings in the CrowdAI dataset, rooms in a 2D floorplan are adjacent, and the bounding box of a single room may encompass parts from multiple rooms. 
Consequently, the attention space inside the RoI of a single room includes irrelevant points from neighboring rooms, which causes confusion. 
By contrast, RoomFormer applies query attention across the entire image feature map, thereby improving its ability to reason about the relationships between adjacent rooms. 
Hence, to deal with such adjacent objects, we will explore strategies enabling feature interaction between polygons in our future work.

All in all, despite the limitations discussed above, it is noteworthy that RoIPoly archives profound improvement for small and medium-sized building outline extraction and is effectively extended for 2D floorplan reconstruction without any post-processing, demonstrating its generalizability. 
Hence, we believe that our approach is beneficial for a wider application of structured prediction. 

\section{Conclusion}
In this paper, we propose a novel Region-of-Interest (RoI) query-based approach, named RoIPoly, for vectorized building outline extraction. 
The query attention is constrained to the most relevant regions of each polygon to allow vertex interactions within the same polygon.
We propose a logit embedding to facilitate the vertex classification, simultaneously removing the invalid vertices with low classification scores to avoid predicting redundant vertices.
RoIPoly achieves excellent performance on the CrowdAI dataset, outperforming the other models that include post-processing in most MS-COCO evaluation metrics and polygon quality metrics.

\section{Declaration of competing interest}
The authors declare that they have no known competing financial interests or personal relationships that could have appeared to influence the work reported in this paper.

\section{Acknowledgement}
This publication is part of the project ``Learning from old maps to create new ones", with project number 19206 of the Open Technology Programme which is financed by the Dutch Research Council (NWO).

%\appendix

%\section{Sample Appendix Section}
%Lorem ipsum dolor sit amet, consectetur adipiscing elit, sed do eiusmod tempor section \ref{sec:sample1} incididunt ut labore et dolore magna aliqua. Ut enim ad minim veniam, quis nostrud exercitation ullamco laboris nisi ut aliquip ex ea commodo consequat. Duis aute irure dolor in reprehenderit in voluptate velit esse cillum dolore eu fugiat nulla pariatur. Excepteur sint occaecat cupidatat non proident, sunt in culpa qui officia deserunt mollit anim id est laborum.

 \bibliographystyle{elsarticle-num} 
 \bibliography{cas-refs}





% \end{thebibliography}
\end{document}